\documentclass[lettersize,journal]{IEEEtran}
\usepackage{amsmath,amsfonts}
\usepackage{algorithmic}
\usepackage{algorithm}
\usepackage{array}
\usepackage[caption=false,font=normalsize,labelfont=sf,textfont=sf]{subfig}
\usepackage{textcomp}
\usepackage{stfloats}
\usepackage{url}
\usepackage{verbatim}
\usepackage{graphicx}
\usepackage{cite}
\usepackage{hyperref}
\usepackage{makecell}
\usepackage{multirow}
\usepackage{multicol}
\usepackage{booktabs}
\begin{document}

\title{DOZE: A Dataset for Open-Vocabulary Zero-Shot Object Navigation in Dynamic Environments}

\author{Ji Ma$^{1}$, Hongming Dai$^{2}$, Yao Mu$^{3}$, Pengying Wu$^{1}$, Hao Wang$^{2}$, Xiaowei Chi$^{4}$, Yang Fei$^{1}$,\\ Shanghang Zhang$^{2}$, and Chang Liu$^{1}$ 
\thanks{Manuscript received: February, 28, 2024; Revised April, 25, 2024; Accepted July, 6, 2024.}
\thanks{This paper was recommended for publication by Editor Markus Vincze upon evaluation of the Associate Editor and Reviewers' comments.
This work was supported by Beijing Nova Program (20220484056) and
 the National Natural Science Foundation of China (62203018).}
\thanks{Ji Ma and Hongming Dai contributed equally to this work. All correspondence should be addressed to Chang Liu and Shanghang Zhang.}
\thanks{$^{1}$Ji Ma, Pengying Wu, Yang Fei and Chang Liu are with the Department of Advanced Manufacturing and Robotics, College of Engineering, Peking University, Beijing 100871, China (e-mail: maji@stu.pku.edu.cn; littlefive@stu.pku.edu.cn; a130321@stu.pku.edu.cn; changliucoe@pku.edu.cn). } 
\thanks{$^{2}$Hongming Dai, Hao Wang and Shanghang Zhang are with the State Key Laboratory of Multimedia Information Processing, School of Computer Science, Peking University, Beijing 100871, China (e-mail: daih@u.nus.edu; wanghao\_cic@tju.edu.cn; shanghang@pku.edu.cn).} 
\thanks{$^{3}$Yao Mu is with the Department of Computer Science, The University of Hong Kong, Pokfulam, Hong Kong (e-mail: ymu@cs.hku.hk).} 
\thanks{$^{4}$Xiaowei Chi is with The Hong Kong University of Science and Technology, Clear Water Bay, Kowloon, Hong Kong (e-mail: xiaowei.chi@connect.ust.hk).}
\thanks{Digital Object Identifier (DOI): see top of this page.}
}

\markboth{%
  \parbox{\textwidth}{%
    IEEE Robotics and Automation Letters. Preprint Version. Accepted July, 2024
  }%
}{}

\maketitle

\begin{abstract}
Zero-Shot Object Navigation (ZSON) requires agents to autonomously locate and approach unseen objects in unfamiliar environments and has emerged as a particularly challenging task within the domain of Embodied AI.
Existing datasets for developing ZSON algorithms lack consideration of dynamic obstacles, object attribute diversity, and scene texts, thus exhibiting noticeable discrepancies from real-world situations. 
To address these issues, we propose a Dataset for Open-Vocabulary Zero-Shot Object Navigation in Dynamic Environments (DOZE) that comprises ten high-fidelity 3D scenes with over 18k tasks, aiming to mimic complex, dynamic real-world scenarios. 
Specifically, DOZE scenes feature multiple moving humanoid obstacles, a wide array of open-vocabulary objects, diverse distinct-attribute objects, and valuable textual hints. 
Besides, different from existing datasets that only provide collision checking between the agent and static obstacles, we enhance DOZE by integrating capabilities for detecting collisions between the agent and moving obstacles. 
This novel functionality enables the evaluation of the agents' collision avoidance abilities in dynamic environments. 
We test four representative ZSON methods on DOZE, revealing substantial room for improvement in existing approaches concerning navigation efficiency, safety, and object recognition accuracy. Our dataset can be found at https://DOZE-Dataset.github.io/.
\end{abstract}

\begin{IEEEkeywords}
Data sets for robot learning, data sets for robotic vision, zero-shot object navigation, semantic scene understanding, embodied AI.
\end{IEEEkeywords}

\section{Introduction}
\label{sec:intro}
\IEEEPARstart{E}{mbodied} AI has made significant progress in recent years, with AI agents gaining enhanced expertise in various tasks such as searching for objects \cite{chaplot2020object}, following language instructions \cite{wu2024camon}, and rearranging objects in realistic scenes \cite{batra2020rearrangement}. 
A key factor behind these achievements is the development of realistic simulation scenarios that simulate high-fidelity real-world environments, offering a systematic and scalable framework for training and evaluating AI agents. 

ObjectGoal navigation (ObjectNav) is a central task in the domain of Embodied AI that showcases the combination of perception, cognition, and action. 
In ObjectNav, an AI agent is required to autonomously navigate an unknown environment and locate a specific object without prior knowledge of the object's location, relying solely on the agent's sensory input and understanding of the environment. 
Building on traditional ObjectNav, where the agent can be pre-trained to recognize the objects it needs to find, Zero-Shot Object Navigation (ZSON) introduces a new challenge where the agent has to search for objects belonging to the categories on which the agent has not been specifically trained. 
Due to the scarcity of dedicated ZSON datasets, researchers predominantly utilize ObjectNav datasets for assessing the efficacy of ZSON methods. 
Existing ObjectNav datasets are composed of either scanned or synthetic 3D scenes. 
The scanned 3D datasets \cite{chang2017matterport3d, ramakrishnan2021habitat} can effectively simulate the diversity of real-world objects and the intricacy of house layouts.
However, the scenes in these datasets often exhibit inaccuracies and distortions in surface textures, are hard to scale, and difficult to modify the contained objects.

To fix these drawbacks, synthetic 3D scene datasets \cite{deitke2020robothor, khanna2023habitat} have been actively developed by emulating real-world settings through the integration of human-authored 3D objects. 
These datasets provide improved scalability and flexibility, allowing for the creation of a wide range of environments with high geometric and textural accuracy.
However, existing synthetic 3D datasets primarily comprise static scenes, thereby failing to accurately reflect the dynamics and stochasticity of real-world scenarios. 
For instance, RoboTHOR \cite{deitke2020robothor} accurately replicates the layout of items in realistic room settings, but fails to account for the vital aspect of moving objects, such as humans, which are essential dynamics within domestic environments.
HSSD-200 \cite{khanna2023habitat} is the latest synthetic 3D scene dataset, which contains moving objects to enable performant embodied AI experiments. Yet, no moving objects are incorporated into the associated ObjectNav tasks.

In addition to the lack of dynamic scenarios, existing ObjectNav datasets are limited by their deficiency in object diversity, notably the scarcity of uncommon objects. 
Such scarcity inevitably restricts perception models to a limited array of items, hindering their ability to recognize and localize objects outside their training data. 
Such shortfall presents a significant challenge in detecting open-vocabulary objects, where perception models must recognize and localize objects across a broad and unspecified vocabulary, including those not encountered during training. 
This capability is crucial because, in practical applications, systems often encounter objects not included in their training datasets. 
Compared to the abundance of open-vocabulary objects in 2D image datasets, the lack of open-vocabulary objects in 3D scene ObjectNav datasets is a notable limitation. 

Another limitation of current ObjectNav datasets is their inability to capture the object characteristics of real-world environments, including spatial and appearance attributes.  In the real world, objects are typically not found in isolation but are related to other objects. For example, pillows are commonly found nestled on sofas, whereas basketballs are usually placed on the floor and may often roll under a chair. Moreover, in a real household, objects within the same category typically exist in multiples and exhibit diverse appearances, such as cups in different colors. 
By including distinct-attribute objects (objects varying in spatial or appearance attributes) in the scenes, it becomes possible to validate ZSON methods' capability to understand the spatial relationship between objects and distinguish similar objects with different appearances.

\begin{figure*}[htbp]
    \centering
    \includegraphics[width=0.75\linewidth]{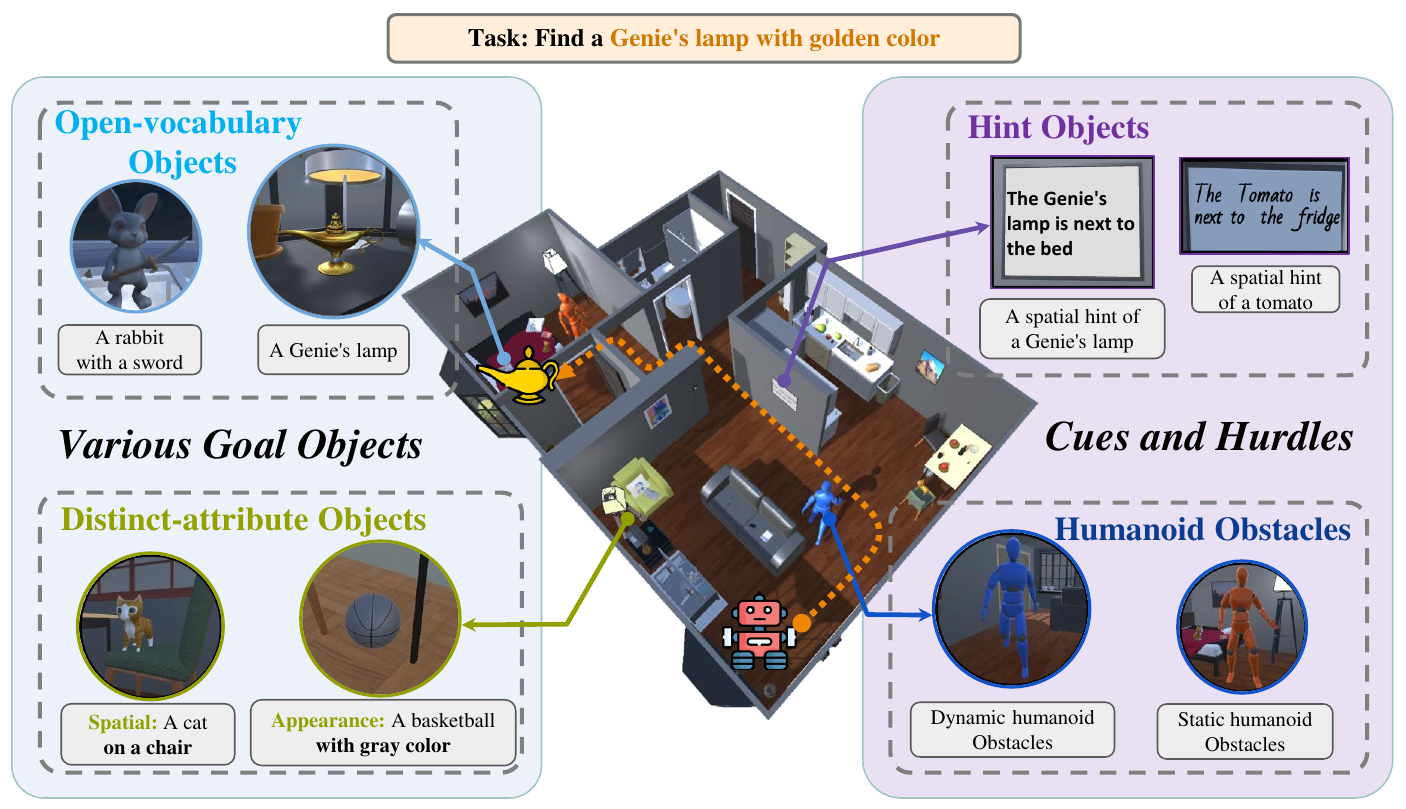}
    \caption{We contribute DOZE: a Dataset for Open-Vocabulary Zero-Shot Object Navigation in Dynamic Environments. This figure shows the task ``Find a Genie's lamp with golden color", and the orange dotted line represents the agent's navigation trajectory. In DOZE, the goal objects include open-vocabulary objects and objects with distinct spatial and appearance attributes. During the navigation process, the agent may encounter boards with textual hints of the target object and humanoid obstacles along the way.}
    \label{top}
    
\end{figure*}

Indoor environments are often rich in text information, including message boards, sticky notes, and to-do lists, all of which, if effectively leveraged, can serve as pivotal contextual clues for object localization, enhancing the capability of agents to identify goal objects within text-rich scenes. 
However, existing datasets often contain no textual hints for object localization, highlighting the urgent need for the creation of datasets that incorporate such hints. 
The inclusion of hint texts poses a unique challenge to traditional ZSON methods, which rely solely on visual perception. 
The advent of datasets augmented with textual information demands innovative methods with a multimodal understanding capability, enabling agents to improve their proficiency in locating goal objects through the integration of both visual and textual cues.

To address existing drawbacks in ObjectNav datasets, we present a novel dataset called DOZE (a \textbf{D}ataset for \textbf{O}pen-vocabulary \textbf{Z}ero-shot object navigation in dynamic \textbf{E}nvironments), which includes ten high-fidelity synthetic 3D scenes and over 18k tasks. 
Differing from existing ObjectNav datasets \cite{chang2017matterport3d, ramakrishnan2021habitat, kolve2017ai2, khanna2023habitat, deitke2020robothor, szot2021habitat}, DOZE stands out by incorporating dynamic obstacles (humanoids), open-vocabulary objects, objects with distinct spatial and appearance attributes, and hint objects. 
Additionally, in contrast to previous datasets that are limited to detecting collisions between the agent and static obstacles, we offer an interface specifically designed to detect collisions between agents and humanoid obstacles, thereby enhancing the assessment of agents' abilities to avoid collisions.
We have tested four representative ZSON methods and found that DOZE presents substantial challenges to these methods, highlighting the considerable potential for enhancing navigation efficiency, safety, and object recognition accuracy. 
Furthermore, we pioneer the use of hint information in ZSON methods by introducing a hint-assisted navigation method. This innovative method empowers agents to locate goal objects more swiftly and accurately by leveraging hint objects, opening up a new research direction for ZSON tasks.

\section{Related Work}

\subsection{Zero-Shot Object Navigation (ZSON) Task}
There has been an increased interest in ZSON tasks \cite{majumdar2022zson, gadre2023cows} in recent years, where agents are evaluated on finding goal objects not included in the training set. 
Compared to traditional ObjectNav tasks that are limited to a fixed set of categories of the goal objects, ZSON allows for a wider range of objects to serve as goals. 
The advent of large language models has greatly enhanced the feasibility of ZSON.
For example, recent language-driven object navigation and task planning methods \cite{dai2023optimal, lian2024tdanet} have advanced the capabilities of agents in understanding and executing complex tasks based on natural language instructions.
Additionally, several studies \cite{shah2023navigation, zhou2023navgpt, wu2024voronav} further demonstrate that agents can be evaluated directly on new object categories in a zero-shot manner. 
Therefore, a dataset that includes a diverse range of goal objects, especially open-vocabulary objects, is in great need.

\begin{figure*}[!ht]
  \centering
  \setlength{\abovecaptionskip}{0cm}
   \includegraphics[width=1\linewidth]{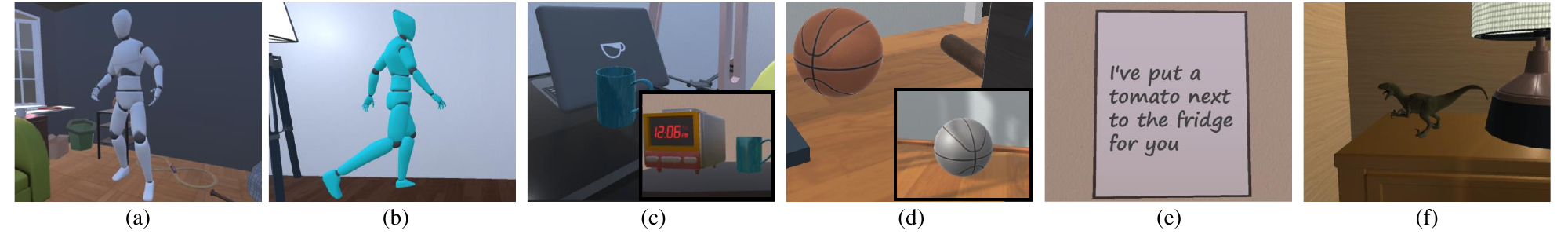}
   {\caption{{Example items in DOZE.} (a) A static humanoid obstacle that obscures a basketball. (b) A dynamic humanoid obstacle walking on the floor. (c) Two mugs with distinct spatial attributes: the left mug is next to a laptop, while the right one is next to an alarm clock. (d) Two basketballs with distinct appearance attributes: the orange basketball on the left and the gray one on the right. (e) A hint whiteboard indicating the location of a tomato. (f) An open-vocabulary object ``Stegosaurus model".}}
   \label{fig:scenes}
   \vspace{-5pt}
\end{figure*}

\vspace{-5pt}

\subsection{ObjectNav Datasets}
Numerous datasets have been utilized in ObjectNav research, including digitally reconstructed scenes derived from environmental scans \cite{chang2017matterport3d, ramakrishnan2021habitat } and synthetic datasets developed within simulation engines \cite{ kolve2017ai2, deitke2020robothor, khanna2023habitat}.
While most ObjectNav datasets present static environments, real-world scenarios often feature dynamic elements, such as moving humans, which can disrupt agent navigation.
Previous research has highlighted the importance of limited sensing field of view and obstacle occlusion consideration in navigation tasks \cite{zhou2024aspire, gao2023probabilistic}.
A recent study \cite{yokoyama2022benchmarking} showed that incorporating dynamic obstacles like moving humans into the training phase of tasks to reach a specific coordinate location, significantly enhances agent generalization, resulting in a higher success rate compared to baseline methods.
Additionally, Khanna et al. \cite{khanna2023habitat} introduced the Habitat dataset, which encompasses social navigation tasks and incorporates moving pedestrians to enhance the complexity of these tasks.
Despite these advancements, a dataset specifically designed for ZSON tasks with dynamic obstacles has been lacking. 
To fill this gap, we present a 3D scene dataset that incorporates both static and dynamic humanoid obstacles and propose the \textit{Collision Rate} (CR) metric, offering new challenges for agent navigation.

Traditional ObjectNav datasets typically designate object categories as goals (e.g., a cup or chair), overlooking the real-world scenario where multiple objects of the same category may exist, each with distinct appearances or spatial relationships. 
In fact, the task of finding goal objects in these scenarios, such as ``find a laptop next to the TV'' or ``find a blue cup'', can pose significant challenges to existing ZSON algorithms, yet there is no dataset that can evaluate such tasks.
Gadre et al. \cite{gadre2023cows} proposed the PASTURE evaluation benchmark that builds on RoboTHOR \cite{deitke2020robothor} validation scenes. Despite the fact that PASTURE focuses on the spatial and appearance attributes of goal objects, the limitation of appearance descriptors to only color and material does not cover a wider range of visual characteristics. 
However, in contrast to PASTURE, our dataset includes objects offering a more detailed and diverse description of appearance attributes, encompassing color, material, texture, and brand.

\section{A Dataset for Open-Vocabulary Zero-Shot
Object Navigation in Dynamic Environments}
\label{sec:formatting}

\subsection{DOZE Overview}
We utilize several scenes from the AI2-THOR \cite{kolve2017ai2} as the basis for DOZE. 
We augment these scenes with a diverse range of objects, including open-vocabulary objects, objects with distinct attributes, and objects with navigation hints.

\begin{figure}[!t]
  \centering
    \includegraphics[width=0.85\linewidth]{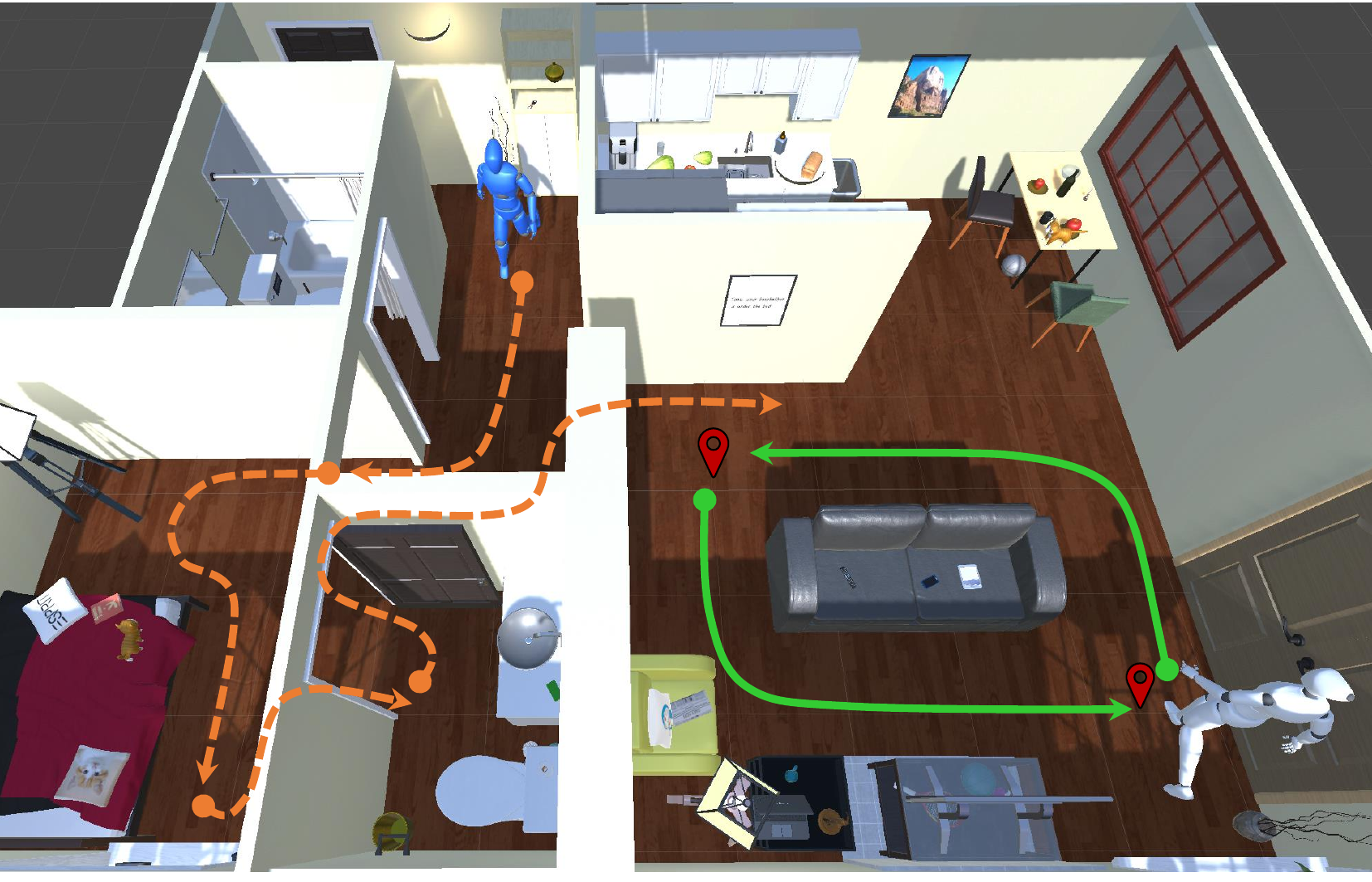}
   \caption{Two motion patterns of dynamic humanoid obstacles. (1) Walking along fixed paths (green solid line) and (2) walking along random paths (orange dashed line).}
   \label{fig:Dynamic human trajectories}
   \vspace{-8pt}
\end{figure}

To mimic the dynamics of real-world indoor environments, we introduce humanoid obstacles that exhibit human-like movements and postures. 
These obstacles include both static humanoid obstacles and dynamic humanoid obstacles that navigate fixed or random paths. 
Consequently, these obstacles are classified into three complexity levels within ZSON tasks, enriching the dataset's realism and challenge:
\begin{itemize}
\item Level 1: Static humanoids obstruct objects.
\item Level 2: Dynamic humanoids move along fixed paths.
\item Level 3: Dynamic humanoids move along random paths. 
\end{itemize}

By combining three levels of humanoid obstacles with four distinct types of goal objects, i.e., open-vocabulary goal objects, goal objects with spatial attributes, goal objects with appearance attributes and goal objects with navigation hints, we create 12 varied ZSON tasks that differ in navigation difficulty and goal object characteristics.
Within each complexity level, the dataset comprises 8,263 tasks emphasizing spatial attributes, 8,814 tasks focusing on appearance attributes, 1,050 tasks related to open-vocabulary goal objects, and 34 tasks incorporating navigation hints.

\vspace{-5pt}

\subsection{Element Design}
\subsubsection{Humanoid Obstacles}
To simulate potential occlusions of goal objects and dynamic obstacles, we incorporate humanoid obstacles into scenes with two distinct behavioral patterns: idling and walking.

Each idling humanoid obstacle is placed around the goal objects to simulate scenarios where the agents' field of view may be partially obstructed by humans, thereby challenging the agents' capabilities of visual detection and active planning.

Each dynamic humanoid obstacle exhibits one of two distinct movement patterns. 
As depicted in Figure \ref{fig:Dynamic human trajectories}, the first pattern involves dynamic humanoids navigating predefined paths, accompanied by intermittent pausing and lingering, while the second pattern entails humanoids moving toward a randomly chosen point within specified areas. 
These dynamic elements enhance the realism of the navigation scenarios and facilitate the evaluation of predictive and planning capabilities of ZSON approaches in dynamic scenes. 
Specifically, the behavior of humanoid obstacles with fixed trajectories can be easily predicted, providing a testing platform for algorithms with motion prediction capabilities. 
In contrast, humanoids with random trajectories are more unpredictable, which challenges agents' path planning capability, requiring immediate reactive path adjustment to avoid collision. 
To the best of the authors' knowledge, our dataset is the first ObjectNav dataset that integrates moving objects in the scene.

\begin{figure}[!t]
  \centering
  \setlength{\abovecaptionskip}{0.cm}
   \includegraphics[width=1\linewidth]{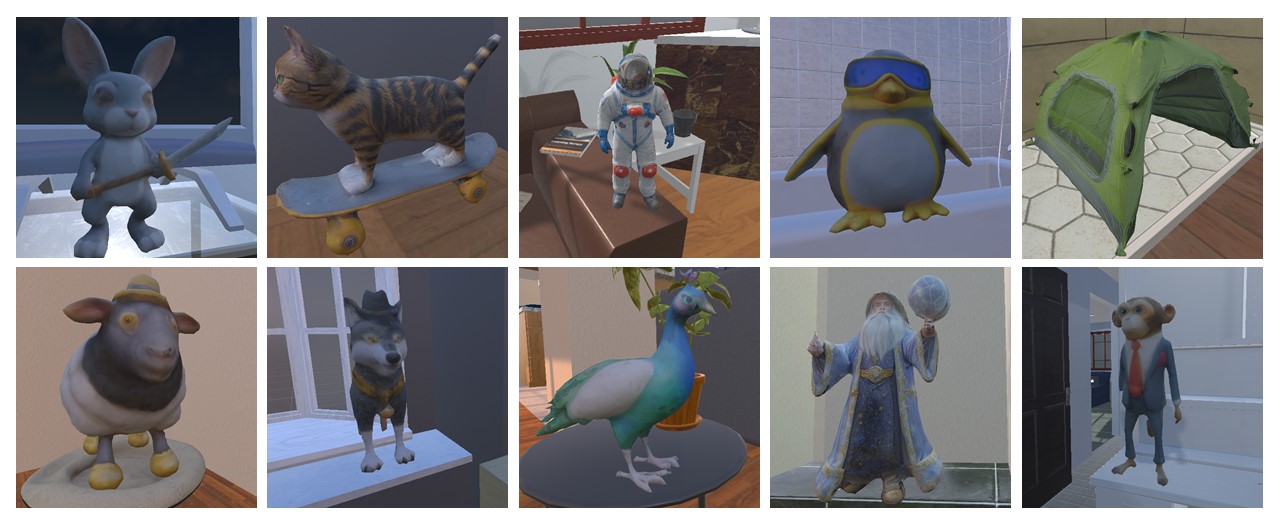}
   \caption{Example open-vocabulary objects from DOZE. Top row: a rabbit holding a sword, a cat on a skateboard, an astronaut model, a penguin wearing goggles, and an outdoor tent. Bottom row: a sheep with a hat, a wolf with a hat, a colorful peacock, a wizard, a monkey in a suit.}
   \label{fig:all_ov}
   \vspace{-3pt}
\end{figure}

\subsubsection{Open-Vocabulary Objects}
We leverage both publicly available free 3D models and Text-to-3D model generation techniques \cite{wang2023prolificdreamer, poole2022dreamfusion} to enrich scenes with a wide variety of open-vocabulary objects, as illustrated in Figure \ref{fig:all_ov}. 
It is noteworthy that our open-vocabulary objects include not only new categories of objects but also objects with enriched properties such as colors, textures, and other characteristics.
Compared to scenes with only a limited, closed set of traditional indoor object categories, the introduction of these open-vocabulary objects presents a more challenging task.
Training on these tasks improves models' ability to generalize in object detection and scene understanding, thereby enhancing agents' adaptability to the diverse and unpredictable objects in the real world.

\subsubsection{Distinct-Attribute Objects}
To assess the ZSON methods' proficiency in reasoning spatial positions and appearance details, we introduce varied spatial and appearance attributes to objects within identical categories as explained below.
\begin{itemize}
\item Spatial attributes. 
By situating objects in specific locations relative to other items, we create goal objects that require understanding spatial relationships, such as ``a pillow on a sofa'' and ``a garbage can next to a stool.'' 
The design of such objects aims to test the agents' spatial reasoning ability in complex environments.
\item Appearance attributes. 
We enhance object appearance variety through attributes like color, material, texture, and pattern, such as ``a gray basketball" vs ``an orange basketball", ``a GROHE toilet" vs ``a TOTO toilet", and ``a pillow with cat images" vs ``a pillow with deer pattern'' in DOZE.
These objects serve to assess the agents' abilities to recognize and understand the visual characteristics of indoor objects.
\end{itemize}
Figure \ref{fig:Attributes} shows the distribution of objects with unique attributes.

\begin{table*}
  \begin{center}
    \setlength{\abovecaptionskip}{0.cm}
      \caption{ObjectGoal Navigation Dataset Statistics}
      \label{tab:ObjectGoal navigation dataset statistics}
        \begin{tabular}{ccccccccc}
        \toprule
         { \multirow{2}{*}{\raisebox{-0.7mm}[0pt][0pt]{Dataset}}} & \multicolumn{2}{c}{Total} & \multicolumn{2}{c}{Average per scene} &
         {\multirow{2}{*}{\makecell[c]{\raisebox{-0.7mm}[0pt][0pt]{Dynamic obstacles}\\\raisebox{-0.7mm}[0pt][0pt]{(\textit{Fixed}$\,$/$\,$\textit{Random})}}}} & {\multirow{2}{*}{\makecell[c]{\raisebox{-0.7mm}[0pt][0pt]{OV objects }}}} & \multirow{2}{*}{{\makecell[c]{\raisebox{-0.7mm}[0pt][0pt]{Distinct-attribute objects}\\\raisebox{-0.7mm}[0pt][0pt]{(\textit{Spacial}$\,$/$\,$\textit{Appearance})}}}} & \multirow{2}{*}{{\raisebox{-0.7mm}[0pt][0pt]{Hint objects}}}\\
         \cmidrule(r){2-3}
         \cmidrule(r){4-5}
         &  Scenes & Objects & Nav. area  & Nav. comp. &  &  &\\ 
        \midrule
        \addlinespace[1mm]
        Replica &  18 & - & 31.11 & 5.99 & $\times$ & $\times$ & $\times$ & $\times$ \\
        \addlinespace[1mm]
        ReplicaCAD & 90 & 92 & 49.8 & 7.2 & $\times$ & $\times$ & $\times$ & $\times$ \\
        \addlinespace[1mm]
        RoboTHOR & 75 & 652 & 25.9 & 2.06 & $\times$ & $\times$ & $\times$ & $\times$ \\
        \addlinespace[1mm]
        Habitat-Matterport 3D & 1000 & 143k & 114.4 & 13.31 & $\times$ & $\times$ & $\times$ & $\times$ \\
        \addlinespace[1mm]
        DOZE (Ours) & 10 & 324 & 46.7 & 4.75 & \makecell[c]{$\checkmark$ \\(5.57$\,$/$\,$4.07)} & \makecell[c]{$\checkmark$ \\(4.8)} & \makecell[c]{$\checkmark$ \\(38.2$\,$/$\,$39.7)} & \makecell[c]{$\checkmark$ \\(3.4) }\\
        \bottomrule
      \end{tabular}
  \end{center}
  \vspace{-15pt}
\end{table*}

\subsubsection{Hint Objects}
\label{sechint}
\begin{figure}[!t]
  \centering
   \includegraphics[width=0.75\linewidth]{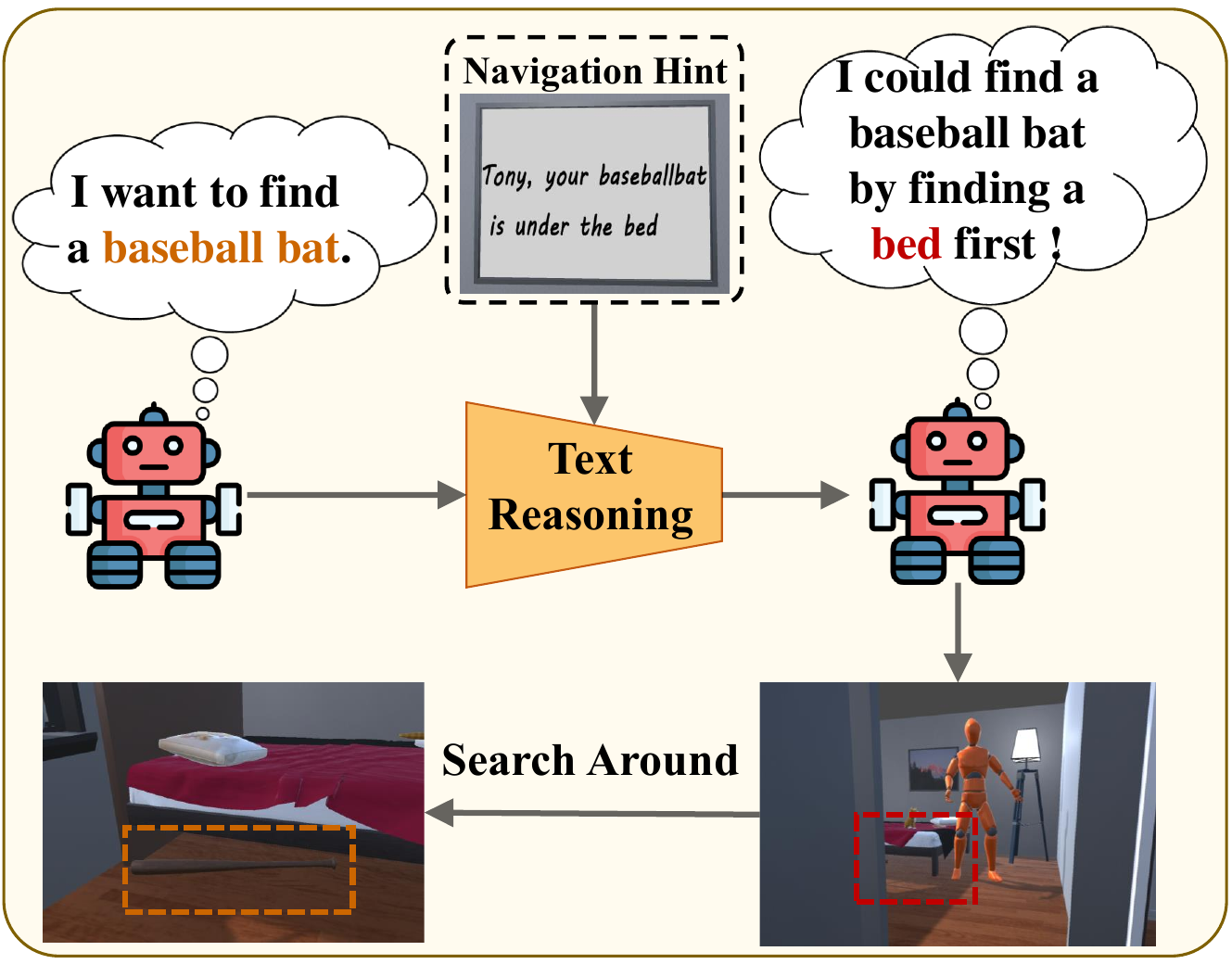}
   \caption{Demonstration of a robot employing text-based reasoning for ZSON tasks. The agent is tasked with seeking a baseball bat (top left) and comes across a navigation hint (top center). Through text reasoning, the robot anticipates finding the bed first (top right). After locating the bed, outlined in a red dashed line (bottom right), the agent explores the surroundings and locates the baseball bat, highlighted with an orange dashed line (bottom left).}
   \label{fig:hint}
   \vspace{-10pt}
\end{figure}

We emphasize the significance of textual information as a crucial source of contextual understanding in diverse indoor settings. 
To enable the agent's ability to deduce the potential location of the goal object from hints during ZSON tasks and expedite the finding of the goal object, we introduce whiteboards with textual content including notes, instructions, reminders, and to-do lists.  
These textual hints assess the agent's ability to interpret and leverage textual information for decision-making.
For instance, if the agent is assigned to find a ``baseball bat'' and encounters a whiteboard stating, ``Tony, your baseball bat is under the bed,'' the agent can then prioritize the bed as an intermediate goal. Upon finding the bed, the agent can then explore the vicinity to ultimately localize the baseball bat, as illustrated in Figure \ref{fig:hint}. 

\begin{figure}[!t]
  \centering
  \setlength{\abovecaptionskip}{0.cm}
    \includegraphics[width=1\linewidth]{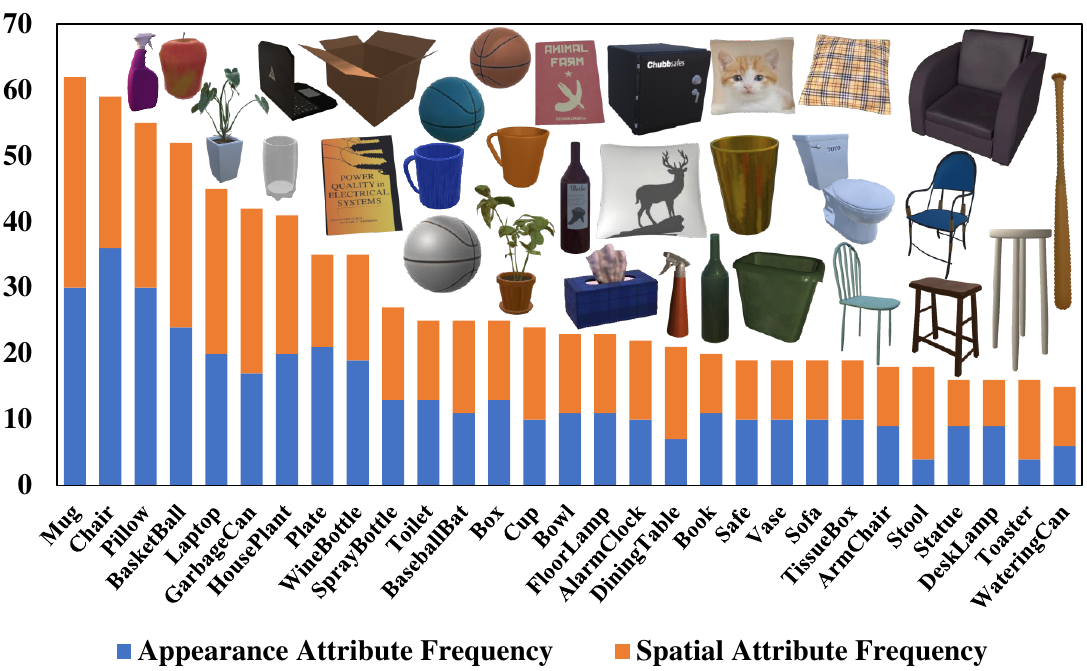}
   \caption{The distribution of distinct-attribute objects and visual examples in DOZE. The distribution highlights the most frequently occurring object categories. 
   Blue bars represent the frequency of objects with appearance attributes, while orange bars indicate the frequency of objects with spatial attributes. The image above the bar chart shows examples of objects with distinct attributes in DOZE.}
   \vspace{-3pt}
   \label{fig:Attributes}
\end{figure}

\subsection{Dataset Statistics}
We compare DOZE with four representative ZSON datasets for scene complexity and object diversity, including the Replica \cite{replica19arxiv}, ReplicaCAD \cite{szot2021habitat}, RoboTHOR \cite{deitke2020robothor}, and Habitat-Matterport 3D \cite{ramakrishnan2021habitat}. 
The comparison, detailed in Table \ref{tab:ObjectGoal navigation dataset statistics}, includes several metrics: ``Scene" for the number of scenes, ``Objects" for the number of unique objects, ``Nav. Area" indicates the navigation area (Average navigable area per scene), and ``Nav. Comp." represents navigation complexity as defined by Ramakrishnan et al. \cite{ramakrishnan2021habitat}. 
``Dynamic obstacles" indicates whether scenes contain dynamic obstacles, ``OV Objects" signifies if open-vocabulary objects are treated as goal objects, ``Distinct-Attribute Objects" denotes if objects of the same category, endowed with different appearance or spatial attributes, are treated as goal objects, and ``Hint Objects" represents if scenes contain textual hints related to the goal objects.

\subsubsection{Scene Complexity}
DOZE contains $10$ scenes, each with an average of $46.7$ square meters of navigable area.
The navigation complexity is quantified by the maximum ratio of the geodesic path distance to the Euclidean distance between any two navigable points \cite{batra2020objectnav}, which reflects the degree of navigation challenges presented by the scene's layout. 
Notably, this metric is specifically designed to assess the complexity of static environments. Consequently, we focus our discussion on the results obtained from level 1 of the DOZE dataset, which exhibits a navigation complexity score of $4.75$, in Table \ref{tab:ObjectGoal navigation dataset statistics}. 
Although this value is lower than those of some other datasets, a distinctive characteristic of DOZE is its inclusion of dynamic obstacles, which notably enhances the navigation complexity.

\subsubsection{Object Diversity} 
Table \ref{tab:ObjectGoal navigation dataset statistics} reveals that, while DOZE has fewer objects than RoboTHOR and Habitat-Matterport 3D, our dataset distinguishes itself through the inclusion of dynamic obstacles, open-vocabulary objects, distinct-attribute objects, and hint objects. 
Dynamic obstacles, a subset of humanoid obstacles, contribute to the complexity of navigation tasks. 
To quantify this complexity, we introduce the \textit{Humanoid Density (HD)} metric, which measures the number of moving humanoids per 100 square meters of navigable space. The average \textit{HD} values are $5.57$ for scenes with fixed trajectory obstacles and $4.07$ for those with random trajectories. 
To enhance the diversity of goal objects in ZSON tasks, DOZE features scenes with an average of $4.8$ diverse open-vocabulary objects, alongside $38.2$ objects with unique locations and $39.7$ with unique appearances. Additionally, it includes about $3.4$ hint objects per scene.

\section{Experiment}
\subsection{Basic Settings} 
\subsubsection{Environment} The ZSON task involves the agent navigating an unknown indoor environment to find an object where the category \( c_i \) of the object is one of the defined categories in the list \( C = \{c_0, \ldots, c_m\} \) with \( m \) indicating the total number of object categories. For each navigation episode, the agent is initialized at a random position \( p_j = \{x_j,y_j,z_j\} \) in the scene \( s_k \). At each timestep \( t \), the agent observes the environment and takes an action \( a_t \). The observation comprises RGB-D frames, the real-time pose of the agent, and the goal object category. The maximum number of steps in an episode is $500$. 

\subsubsection{Agent} The agent used in the experiment is a LoCoBot \cite{locobot} with a height of $0.9m$ and a radius of $0.18m$, equipped with an Intel RealSense RGB-D camera. 
The agent's action space includes the following actions: \texttt{MoveAhead}, \texttt{RotateLeft}, \texttt{RotateRight}, \texttt{Done} with a default forward step of $0.25m$ and a turn angle of 30 degrees. 
An episode is considered successful if the agent takes the \texttt{Done} action when its distance from the goal is within $1m$.
Here, the large distance is utilized because there may be humanoid obstacles obstructing the goal objects in our dataset. 
It is counted as a collision if the agent is within $0.2$ meters of any humanoid obstacle. 

\subsubsection{Navigation Metrics} We use standard ZSON metrics to measure method performance, including
\begin{itemize}
\item Success Rate (SR): the percentage of episodes where the agent executes \texttt{Done} less than $1.0m$ from the target.
\item Success weighted by Path Length (SPL) \cite{anderson2018evaluation}: 
a metric that weights success by the ratio between the shortest path length and the actual path length. This metric measures the efficiency of navigation relative to the shortest paths.
\item Collision Rate (CR): the probability of the agent colliding with humanoid obstacles during navigation, defined as
\begin{equation}
    \setlength{\abovedisplayskip}{3pt}
    \setlength{\belowdisplayskip}{3pt}
     \textit{CR} = \frac1N\sum_{i=1}^N\frac{C_i}{S_i} \label{equation_CR},
\end{equation}
where $N$ is the total number of episodes, the number of collisions in episode $i$ is represented by $C_i$, and the number of steps taken by the agent in this episode is represented by $S_i$. 
Compared to existing ObjectNav benchmarks' focus on task completion, the \textit{CR} metric offers a complementary assessment of agents' safe navigation ability.
\end{itemize}

\begin{table*}[!t]
    \begin{center}
        \caption{Experiment results (SR$\uparrow$ / SPL$\uparrow$ / CR $\downarrow$)}
         \label{tab:result}
         \setlength{\tabcolsep}{4pt}
          \begin{tabular}{cccccccccc}
            \toprule
            \multirow{2}{*}{\raisebox{-0.7mm}[0pt][0pt]{Method}} & \multicolumn{3}{c}{Level 1} & \multicolumn{3}{c}{Level 2} & \multicolumn{3}{c}{Level 3} \\
             \cmidrule(r){2-4}
             \cmidrule(r){5-7}
             \cmidrule(r){8-10}
             &  OV & Spatial & Appearance  &  OV & Spatial & Appearance  &  OV & Spatial & Appearance   \\ 
            \midrule
            \addlinespace[1mm]
            \multirow{1}{*}{Random}
                  & 6.3$\,$/$\,$2.1$\,$/$\,$1.1 & 15.6$\,$/$\,$3.3$\,$/$\,$1.0 & 27.8$\,$/$\,$5.2$\,$/$\,$0.7 & 6.3$\,$/$\,$1.5$\,$/$\,$7.0	& 12.4$\,$/$\,$2.8$\,$/$\,$7.3	& 24.9$\,$/$\,$4.7$\,$/$\,$6.9 &	4.1$\,$/$\,$0.4$\,$/$\,$17.1 &	2.4$\,$/$\,$0.3$\,$/$\,$17.9	& 3.6$\,$/$\,$0.4$\,$/$\,$20.2\\
                  \addlinespace[1mm]
            \multirow{1}{*}{Frontier}
                  & 10.4$\,$/$\,$3.4$\,$/$\,$0.8 &	22.3$\,$/$\,$6.1$\,$/$\,$0.8 & 32.5$\,$/$\,$8.1$\,$/$\,$0.9 & 10.4$\,$/$\,$2.9$\,$/$\,$6.2 & 19.8$\,$/$\,$4.5$\,$/$\,$6.9 & 29.1$\,$/$\,$6.2$\,$/$\,$6.5 & 4.1$\,$/$\,$0.3$\,$/$\,$15.3 & 3.0$\,$/$\,$0.6$\,$/$\,$16.1 & 5.4$\,$/$\,$0.5$\,$/$\,$15.8 \\ 
                  \addlinespace[1mm]
            \multirow{1}{*}{C-L3MVN}
                 & 12.5$\,$/$\,$3.9$\,$/$\,$0.8 & 25.4$\,$/$\,$7.3$\,$/$\,$0.7 & 36.1$\,$/$\,$9.5$\,$/$\,$0.8 & 10.4$\,$/$\,$3.1$\,$/$\,$4.6 & 23.6$\,$/$\,$5.8$\,$/$\,$5.1 & 34.2$\,$/$\,$7.5$\,$/$\,$4.9 & 6.3$\,$/$\,$0.7$\,$/$\,$12.1 & 6.5$\,$/$\,$0.5$\,$/$\,$13.9 & 8.3$\,$/$\,$0.6$\,$/$\,$14.1 \\ 
                 \addlinespace[1mm]
            \multirow{1}{*}{C-LGX}
                   & 10.4$\,$/$\,$3.8$\,$/$\,$0.6 &23.8$\,$/$\,$7.0$\,$/$\,$0.6&32.8$\,$/$\,$9.2$\,$/$\,$0.8&8.3$\,$/$\,$2.9$\,$/$\,$5.8 &18.3$\,$/$\,$4.7$\,$/$\,$5.3&27.2$\,$/$\,$6.7$\,$/$\,$5.3&6.3$\,$/$\,$0.4$\,$/$\,$11.3&7.8$\,$/$\,$0.6$\,$/$\,$14.1&9.2$\,$/$\,$0.7$\,$/$\,$14.2\\ 
            \bottomrule
          \end{tabular}
          
    \end{center}
   \vspace{-10pt}
\end{table*}


\subsection{Three-Level ZSON Experiment }
\subsubsection{Benchmarks} 
To evaluate the influence of DOZE's diverse characteristics on ZSON, we use four representative benchmarks to conduct comprehensive evaluations of DOZE across open-vocabulary goal objects, goal objects with spatial attributes and goal objects with appearance attributes. The experiments regarding goal objects with hints are detailed in Section \ref{Hint-informed ObjectNav Experiment}. 
\begin{itemize}
    \item Random: The method drives the agent to move to randomly chosen points in an unexplored area.
    \item Frontier \cite{yamauchi1997frontier}: This method is based on the frontier map to select points from the closest border points between the unexplored and unoccupied areas as intermediate exploration goals for the agent.
    \item C-L3MVN: L3MVN \cite{yu2023l3mvn} uses the Large Language Model (LLM) for object navigation to improve exploration and search efficiency. To handle episodes involving dynamic humanoid obstacles, we enhance the L3MVN with a collision-avoidance feature through an augmented semantic map. This improvement dynamically incorporates the motion of humanoid robots into the map by continuously detecting ``robots'' in the environment using a visual model. As these robots move, the areas they occupy are identified and marked as unreachable zones at each navigational step. This ensures that the agent avoids these newly forbidden zones, enabling safer exploration.
    \item C-LGX: LGX \cite{dorbala2023can} combines the Vision Language Model and LLM for intuitive, commonsense-driven zero-shot object navigation. We introduce Collision-Aware Language-Guided Exploration (C-LGX) by refining the original LGX. At each time step, C-LGX uses GLIP to detect the goal object and any dynamic humanoid obstacles. It then uses the positions of these obstacles observed over two consecutive steps to predict their future movements and locations. Based on this prediction, the agent navigates in the opposite direction to avoid potential collisions.
\end{itemize}

\begin{figure}[!t]
    \centering
    \setlength{\abovecaptionskip}{0.cm}
    \includegraphics[width=0.475\textwidth, keepaspectratio=false]{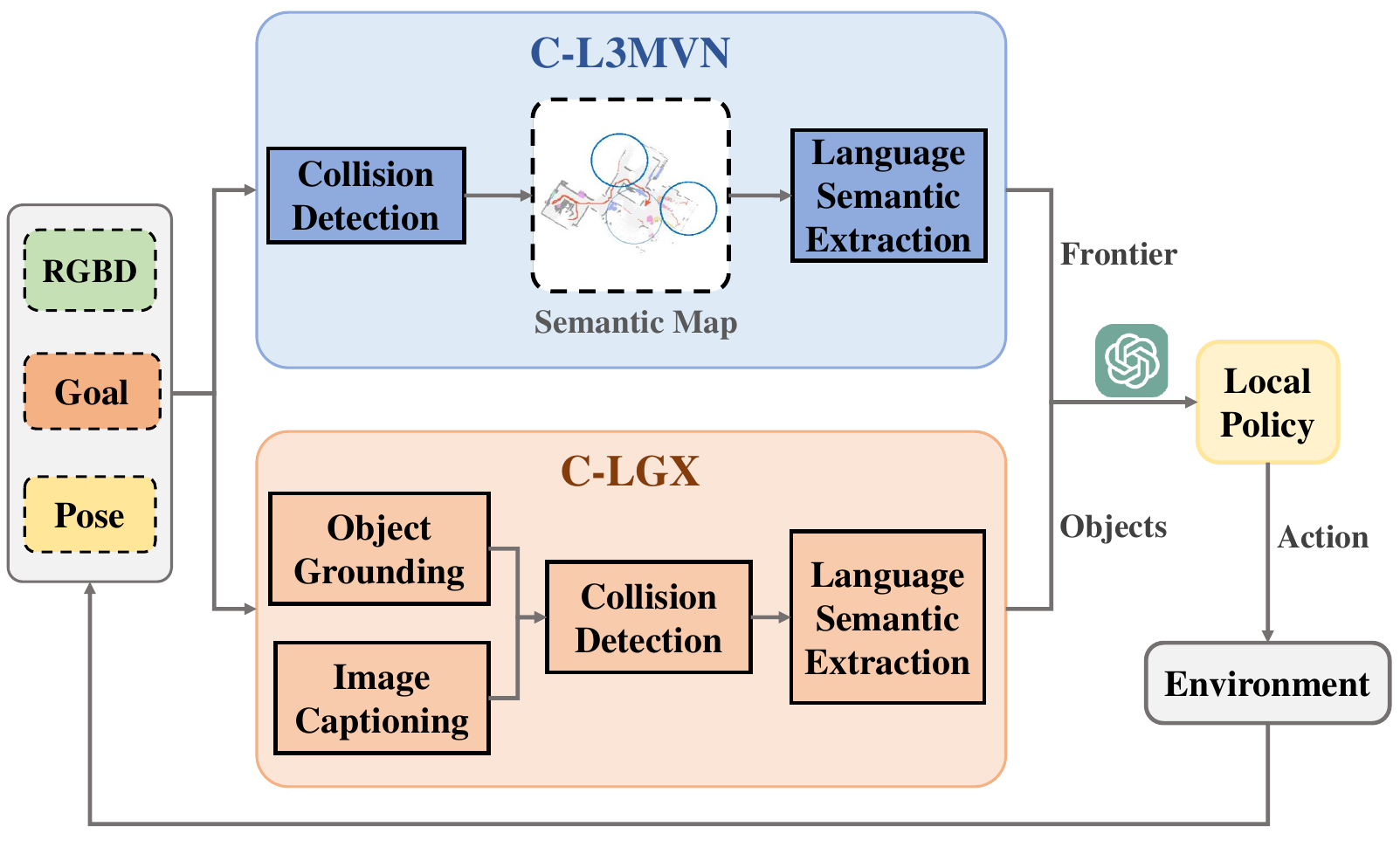}
    \caption{The architecture of the C-L3MVN and C-LGX. 
    The text ``RGBD'' represents the RGB-D observation of the agent, ``Goal'' represents the agent's target object, and ``Pose'' represents the agent's position.}
    \label{fig:Collision-aware Language Guided Exploration Architecture}
    \label{comparison}
    \vspace{-5pt}
\end{figure}

The differences between C-L3MVN and C-LGX are described in Figure \ref{comparison}. 
Both approaches employ the LLM as a decision maker in the exploration module to leverage the common sense reasoning ability of LLM. 
However, these two approaches differ in whether a semantic map is generated. 
In particular, C-L3MVN leverages the rich information from the semantic map and queries the LLM for frontier-based exploration choice. C-LGX, in contrast, solely relies on the LLM for making exploration decisions based on the current observation.

\subsubsection{Level-Specific Experiments} 
We evaluate the four methods on tasks of three levels, each containing three types of goal objects: goal objects with spatial attributes, goal objects with appearance attributes, and open-vocabulary goal objects. In this section, we report and discuss the performance metrics for each type of goal object separately, providing a comprehensive understanding of the results.

As shown in Table \ref{tab:result}, the absence of dynamic humanoid obstacles simplifies the navigation task in Level 1 tasks. The main challenge for the agent is whether the detector can identify the obstructed objects. In this level, C-LGX has achieved an average success rate of 22.3\%, which is marginally better than the Random agent but worse than the Frontier-based agent. It indicates that the mapless exploration strategy of C-LGX is limited even with the help of LLM. C-L3MVN, in contrast, demonstrates the best performance in the task with an average success rate of 24.7\% by integrating the LLM with the semantic map for exploration. As expected, the static humanoid obstacles result in meager collision rates for all four methods on Level 1. 

In Level 2, C-L3MVN maintains its robust navigation performance, with most metrics aligning with those of Level 1. However, the performance of C-LGX demonstrates a notable decline compared to that in Level 1. The decline suggests that the semantic map in C-L3MVN proved effective in avoiding dynamic obstacles, where dynamic humanoid obstacles on fixed trajectories were perceived as continuous obstacles, which are easy for the agent to bypass. In contrast, the C-LGX, which gathers perception of its surroundings by rotating in place at each navigation step, encountered significant difficulties due to dynamic humanoid obstacles whose paths often intersected with the agent's viewing horizon, thus complicating the BLIP's scene understanding. 
The extra overhead brought by the collision avoidance module often results in delays in real-time planning, frequently leading to collision due to delayed reactions. 

\begin{table}[!t]
    \begin{center}
        \caption{Results of DOZE Ablation using C-L3MVN (SR$\uparrow$ / SPL$\uparrow$)}
         \label{tab:ablation_result}
          \begin{tabular}{cccc}
            \toprule
            Objects in scenes & OV & Spatial & Appearance \\
            \midrule
              All types of objects & 12.5$\,$/$\,$3.8 & 23.8$\,$/$\,$7.0 & 32.8$\,$/$\,$9.2 \\
                            Target type objects only & \textbf{14.6$\,$/$\,$4.2} & \textbf{26.5$\,$/$\,$7.7} & \textbf{35.5$\,$/$\,$9.7} \\
            \bottomrule
          \end{tabular}
    \end{center}
\end{table}

Table \ref{tab:result} shows that both C-L3MVN and C-LGX exhibit significant potential for improvement in Level 3. The performance drop from Level 1 to Level 3 in C-L3MVN was primarily attributed to the inadequacy of its semantic mapping module in modeling moving humanoids. Instead of representing the humanoid obstacles as intended moving points, the module erroneously modeled the obstacles as continuous obstacles, severely limiting C-L3MVN's exploration area and leading to numerous failures due to exceeding the maximum exploration steps. C-LGX, by avoiding introducing semantic maps for exploration, managed to bypass this stalling issue. However, the enhancement in performance realized through the application of C-LGX cost maps remains marginal, as per experimental observations. This limitation is attributed to the cost maps' inability to represent the comprehensive environmental information provided by semantic maps. 

We observe a consistent decrease in the SPL and an increase in the CR from Level 1 to Level 3 for all methods. It is therefore evident that existing ZSON methods fall short in scenarios featuring dynamic moving obstacles though demonstrating excellent performance in traditional ZSON tasks.

\vspace{-10pt}

\subsection{Ablation Study on OV and Distinct-Attribute Objects}
Table \ref{tab:result} shows the performance of various ObjectNav methods in finding three different types of goal objects, while all three types of objects exist in the scenes.
To further evaluate the impact of each object type,
we conduct three ablation studies. 
The first one involves OV object-only experiments where only OV objects exist in scenes, and the goal object is also an OV object. 
Similarly, the second and third studies involve spatial attributes-only objects and appearance attributes-only objects, respectively.
We select the best-performing method C-L3MVN for experiments in Level 1 tasks \footnote{As Table \ref{tab:result} shows, Level 2 and 3 tasks are challenging due to the inclusion of moving humanoids, and can obscure effects of OV objects and distinct-attribute objects. Therefore, we only use Level 1 tasks in this ablation study.}.
This design allows us to assess the contribution of each object type to the model's performance in isolation. 
Experimental results presented in Table \ref{tab:ablation_result} reveal performance decrement in scenes when all object types are included compared to scenes restricted to a single object type, suggesting each object type indeed introduces additional challenges to ObjectNav approaches.

\begin{figure}[!t]
    \centering
    \includegraphics[width=0.475\textwidth]{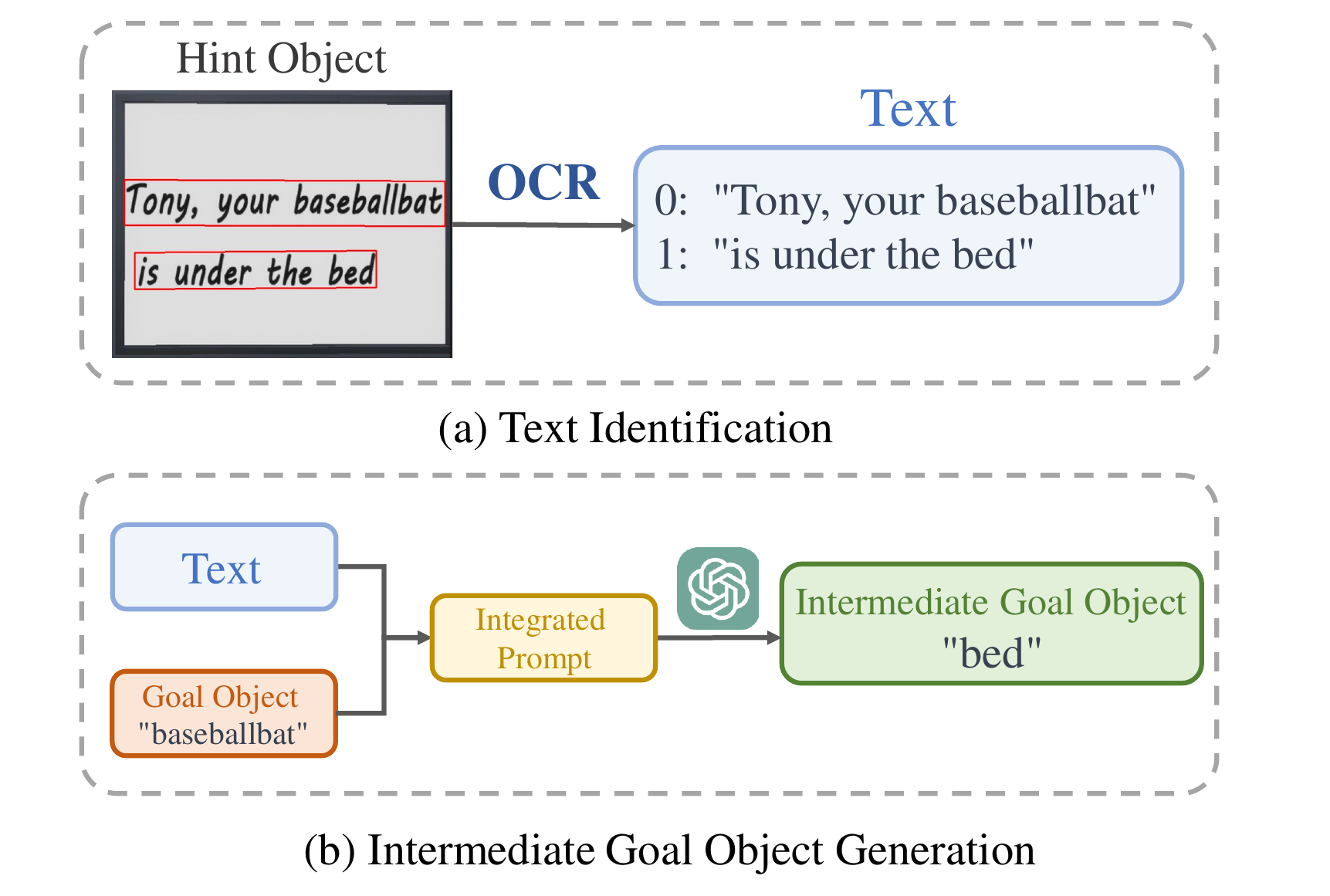}
    \caption{Acquisition of intermediate goal object. (a) The OCR tool is used to recognize the writing on the whiteboard as text. (b) The recognized text is combined with the goal object to form the integrated prompt, which is subsequently transmitted to the LLM to generate the intermediate goal object.}
    \label{fig:Indirect object}
\end{figure}


\subsection{Hint-Informed ZSON Experiment}
\label{Hint-informed ObjectNav Experiment}
The hint-informed ZSON task defines a navigational challenge where the agent actively utilizes environmental hints to acquire crucial information. These hints comprise textual data that pinpoints the object's specific location. 
The agent can use hints to extract intermediate target objects, which are usually easily identifiable and are placed in the vicinity of the goal objects.
Finding the intermediate target objects can make it easier to locate the goal objects. 
To address the new challenge of ZSON with hint information, we augment the Frontier and L3MVN methods to utilize hints in the environment, resulting in H-Frontier and H-L3MVN, respectively.
Specifically, when the agent identifies ``Hint Object" as shown in Figure \ref{fig:Indirect object} during navigation, the agent extracts textual information with the OCR module. Then, the gathered text hints and the goal object are combined and fed into the Large Language Model as an integrated prompt to acquire the intermediate target object. Intermediate target objects are generally easier to recognize and can aid in effectively navigating toward the goal object.
Upon the agent's arrival at the intermediate target object, the agent rotates in place and exhaustively searches the surrounding environment to detect the goal object. 

We evaluate four methods—Frontier, H-Frontier, L3MVN, and H-L3MVN—across three scene levels: Level 1, Level 2, and Level 3, selecting goal objects that each have corresponding hint objects for all experiments.
The results presented in Table \ref{tab:result of hint} show that the two hint-informed methods, H-Frontier and H-L3MVN, consistently exhibit higher success rates compared to their baseline counterparts, Frontier and L3MVN, across all three levels. More notably, there is a significant enhancement in SPL for the hint-informed methods, which suggests that incorporating hint information into the scene can effectively enhance the efficiency of the ZSON agent in locating goal objects. 

\begin{table}[!t]
    \begin{center}
    \setlength{\abovecaptionskip}{0.2cm}
        \caption{Hint-based experiment results (SR$\uparrow$ / SPL$\uparrow$)}
        \label{tab:result of hint}
            \begin{tabular}{ccccccc}
            \toprule
             Method & \multicolumn{1}{c}{Level 1} & \multicolumn{1}{c}{Level 2} & \multicolumn{1}{c}{Level 3} \\
            \midrule
            \multirow{1}{*}{Frontier}
                  & 26.5$\,$/$\,$6.8 & 23.5$\,$/$\,$5.1 & 8.8$\,$/$\,$0.8 \\  
            \multirow{1}{*}{H-Frontier}
                 & 26.5$\,$/$\,$7.5 & 26.5$\,$/$\,$6.2 & 10.3$\,$/$\,$1.4  \\ 
            \multirow{1}{*}{L3MVN}
                  & 32.3$\,$/$\,$9.1 & 29.4$\,$/$\,$7.3 & 8.8$\,$/$\,$0.9 \\  
            \multirow{1}{*}{H-L3MVN}
                 & \textbf{35.2$\,$/$\,$13.3} & \textbf{29.4$\,$/$\,$10.5} & \textbf{11.7$\,$/$\,$1.6}\\ 
            \bottomrule
          \end{tabular}
    \end{center}
    \vspace{-20pt}
\end{table}
\section{Conclusion}
In this paper, we propose DOZE, a dataset for open-vocabulary Zero-Shot Object Navigation in dynamic environments. 
DOZE comprises static and dynamic humanoid obstacles, open-vocabulary objects, objects with distinct spatial and appearance attributes, and hint objects. 
We evaluate the dataset with several representative ZSON methods, and the experimental results highlight the effectiveness of our dataset across multiple facets. 
Moreover, our work introduces a novel method by utilizing hint information during navigation, enabling agents to locate goal objects more swiftly and accurately. 
In the future, we plan to extend the navigation scenarios to encompass large-scale complex outdoor environments and design additional interaction mechanisms between agents and moving humanoids. 

{
    \bibliographystyle{IEEEtran}
    \bibliography{main}
}
\vfill

\end{document}